\documentclass{article}
\usepackage{spconf}
\usepackage{amsmath,graphicx}
\usepackage{times}
\usepackage{epsfig}
\usepackage{graphicx}
\usepackage{amsmath}
\usepackage{amssymb}
\usepackage{array}
\usepackage{multirow}
\usepackage{booktabs}

\title{DSAL-GAN: Denoising based Saliency Prediction with Generative Adversarial Networks}

\address{ $^{\dagger}$ IIIT Sri City, $^{\ddagger}$CSIR-CEERI Pilani, $^{\S}$CCS Computers Pvt Ltd, $^{\$}$IIT Delhi, $^{\P}$IIT Jodhpur .  \and  $^{\dagger}$ prerana.m@iiits.in, $^{\ddagger}$ $\{$mksnith, singhajay518, avinres, akkshitatrivedi$\}$@gmail.com,\and $^{\S}$ meghmak95@gmail.com,$^{\$}$ brejesh@ee.iitd.ac.in, $^{\P}$ schaudhury@gmail.com.  }

\makeatletter
\def\@name{ \emph{Prerana Mukherjee$^{\dagger*}$\thanks{*Equal Contribution} and  Manoj Sharma$^{\ddagger*}$ 
\quad  Megh Makwana$^{\S*}$ \qquad  Ajay Pratap Singh$^{\ddagger}$},  \\ \emph{Avinash Upadhyay$^{\ddagger}$ \qquad Akkshita Trivedi$^{\ddagger}$ \qquad Brejesh Lall$^{\$}$ \qquad Santanu Chaudhury$^{\P}$}}
\makeatother

\begin{document}
%

\maketitle
\begin{abstract}
Synthesizing high quality saliency maps from noisy images is a challenging problem in computer vision and has many practical applications. Samples generated by existing techniques for saliency detection cannot handle the noise perturbations smoothly and fail to delineate the salient objects present in the given scene. In this paper, we present a novel end-to-end coupled Denoising based Saliency Prediction with Generative Adversarial Network (DSAL-GAN) framework to address the problem of salient object detection in noisy images. DSAL-GAN consists of two generative adversarial-networks (GAN) trained end-to-end to perform denoising and saliency prediction altogether in a holistic manner. The first GAN consists of a generator which denoises the noisy input image, and in the discriminator counterpart we check whether the output is a denoised image or ground truth original image. The second GAN predicts the saliency maps from raw pixels of the input denoised image using a data-driven metric based on saliency prediction method with adversarial loss. Cycle consistency loss is also incorporated to further improve salient region prediction. We demonstrate with comprehensive evaluation that the proposed framework outperforms several baseline saliency models on various performance benchmarks.

\end{abstract}

\section{Introduction}
\label{sec:intro}

Selective attentional processing involves suitable processing of visual stimuli to localize the salient objects in the image and hence it is an important area of research.
\begin{figure}[thpb]
       \centering
       \fbox{
       \includegraphics[scale=0.35]{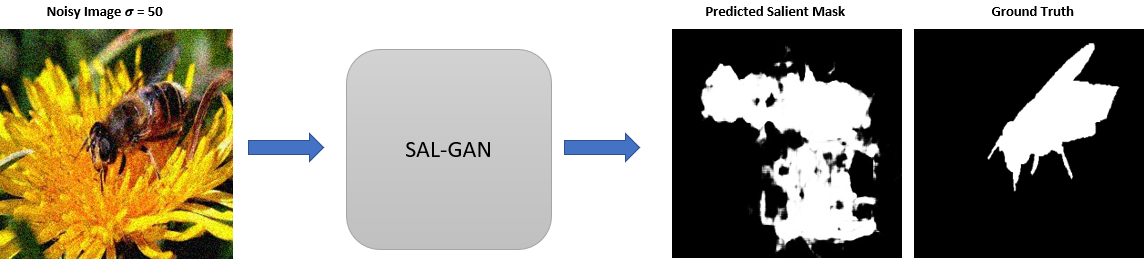}}
       \caption{Resuts of SAL-GAN (trained on clean data for saliency prediction) for noisy image as Input}
       \label{fig:noise}
       \vspace{-0.25in}
\end{figure}
The concept is highly inspired by the inherent working of the human visual system (HVS). In the visual cortex, every neuron responds to a particular section of the visual field. The receptive field (RF) is the area which is responsible for the perception of the visual stimuli. It responds to the center-surround difference detection corresponding to the object's edges. As we progressively ascend to the higher levels in the visual cortex, the object representation is laid out in a hierarchical topography. This selective attentional processing or commonly termed as saliency prediction draws huge attention in the vision community due to its  widespread applicability in various research domain areas such as adaptive image compression \cite{park2017transfer, li2018closed}, video summarization \cite{mademlis2018regularized}, image retargeting\cite{zhou2018perceptually}  etc.
However, when the input image is distorted due to noise perturbations, the detection of the salient object becomes challenging as most of the saliency detection approaches \cite{pan2017salgan} may fail to recognize the salient object in presence of noise and misclassify noise pixels as object pixels as shown in Fig. \ref{fig:noise}.

\begin{figure*}[thpb]
       \centering
       \fbox{
       \includegraphics[scale=0.55]{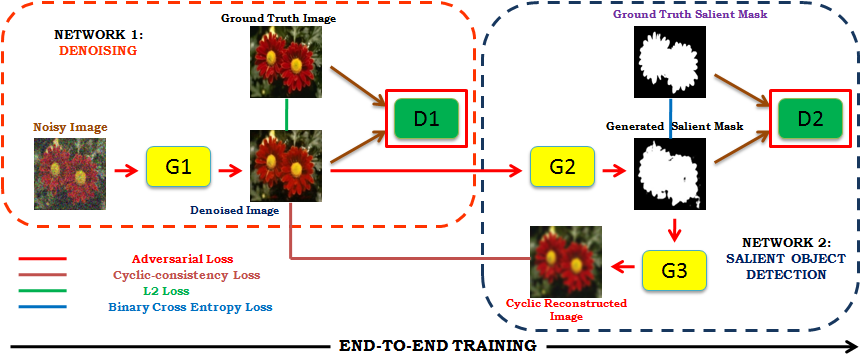}}
       \caption{End-to-End DSAL-GAN Network Architecture.}
       \label{fig:workflow}
\end{figure*}

The closest work to ours is \textit{SalGAN} which estimates the saliency map of an input
image using a deep convolutional neural network (DCNN) utilizing a binary cross entropy (BCE) loss getting propagated across successive down-sampled saliency maps. Furtheron, the model is refined with the discriminator block which aligns the fake (generated saliency output) close to the real (ground truth saliency map) one. However, in presence of noise variations it is not able to handle the salient object prediction as shown in Fig. \ref{fig:noise}. To circumvent this problem, in this work we propose a novel end-to-end coupled Denoising based Saliency Prediction with Generative Adversarial Network (DSAL-GAN) framework. DSAL-GAN consists of coupled dual step generative adversarial network: i) In first generator step, we perform denoising of the input noisy image, and in the discriminator counterpart we check whether the output is a denoised image or ground truth original image, ii) In the second generator step, we predict saliency maps from raw pixels of an input denoised image, and in the discriminator counterpart we discriminate whether a saliency map is a predicted one or ground truth. Due to joint optimisation (end-to-end) training, DSAL-GAN generates the predicted saliency map which is indistinguishable with the ground truth and is capable of handling noisy images in a holistic manner.


In view of above discussions, the key contributions of this paper are:
\begin{enumerate}
    \item Joint optimization of denoising and saliency prediction in a coupled end-to-end trainable GAN framework.
    \item Use of cycle consistency loss to refine saliency prediction.
    \item Exhaustive comparative analysis with several saliency baselines to demonstrate superior performance over various benchmark datasets.
\end{enumerate}

Remaining sections in the paper are organized as follows. In Sec. \ref{sec:methodology}, we outline the methodology we propose to provide a holistic framework for denoising and saliency prediction. In Sec. \ref{sec:results}, we discuss experimental results and conclude the paper in Sec. \ref{sec:conclusion}.

\begin{figure}[thpb]
       \centering
       \fbox{
       \includegraphics[scale=0.5]{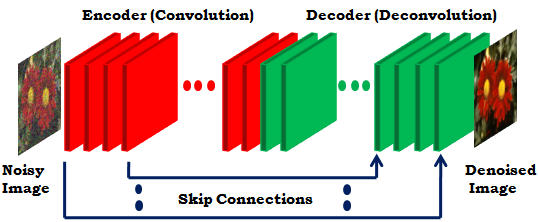}}
       \caption{RED-Net Generator Network Architecture \cite{DBLP:journals/corr/MaoSY16a}.}
       \label{fig:rednet}
       \vspace{-0.25in}
\end{figure}

\section{Methodology}
\label{sec:methodology}

\subsection{Problem Formation}

The problem of denoising can mathematically be formulated as $x = D.y$. The $x$ and $y$ denote the noisy and clean image (without noise perturbations) and $D$  represents the noise matrix which degrades the clean image. Clean image $y$ can be obtained by taking inverse of the noise matrix. The formulation to obtain denoised image $y$ is: 
\begin{equation}
    y = D^{-1}.x = f_d(x)
\end{equation}
Where $f_d$ is denoising function. The saliency map can be obtained using the following equation:
\begin{equation}
         z=f_{SOD}(f_d(x))=f_{SOD}(y)
\end{equation}
Where $z$ is the saliency map of clean image $y$ obtained using denoising function on $x$. The $f_{SOD}$ function is responsible for localization of saliency map.

\subsection{Image Denoising}
As shown in the Fig. \ref{fig:workflow}, the denoising network maps the noisy image $x$ to a clean (denoised) image $y$. The generator $G1$ learns to generate image $y_{predicted}$ from input noisy image $x$, while the discriminator network $D1$ learns to differentiate between $y_{predicted}=y_{p}$ and $y$ (Ground Truth). Here, the generator architecture is similar as given in \cite{DBLP:journals/corr/MaoSY16a}, and shown in Fig. \ref{fig:rednet}. The content loss (L2 Loss) of the generator can be represented as,
\begin{equation}
L_{Content}^{Denoising}=\frac{1}{n}\sum_{i=1}^{N}\parallel G1(x_i) - Y_i) \parallel_2 
\end{equation}
The adversarial loss can be formulated as,
\begin{equation}
L_{Adversarial}^{Denoising}=\frac{1}{n}\sum_{i=1}^{N} -log\,D1(G1(x_i)) 
\end{equation}
Total loss for the denoising network is calculated as,
\begin{equation}
L^{Denoising}=L_{Content}^{Denoising}+w_{1}.L_{Adversarial}^{Denoising}
\end{equation}

\begin{table}[]
\centering
\caption{Discriminator Network Architecture used for both Denoising and Salient Object Detection. s and p denote stride and padding respectively.}
\label{tab:tab1}
\vspace{0.1in}
\scalebox{1}{
\begin{tabular}{ll}
\hline
layers        & Image                                                                                                         \\ \hline
{[}layer 1{]} & \begin{tabular}[c]{@{}l@{}}conv1\_a (1,1,3), s=1, p=1; ReLU\\ conv1\_b (3,3,32), s=1, p=1; ReLU\end{tabular}  \\ \hline
              & pool1 (2,2), s=2, p=0                                                                                         \\ \hline
{[}layer 2{]} & \begin{tabular}[c]{@{}l@{}}conv2\_a (3,3,64), s=1, p=1; ReLU\\ conv2\_b (3,3,64), s=1, p=1; ReLU\end{tabular} \\ \hline
              & pool2 (2,2), s=2, p=0                                                                                         \\ \hline
{[}layer 3{]} & \begin{tabular}[c]{@{}l@{}}conv3\_a (3,3,64), s=1, p=1; ReLU\\ conv3\_b (3,3,64), s=1, p=1; ReLU\end{tabular} \\ \hline
              & pool3 (2,2), s=2, p=0                                                                                         \\ \hline
{[}layer 4{]} & fc4 (100); tanh                                                                                               \\ \hline
{[}layer 5{]} & fc5 (2); tanh                                                                                                 \\ \hline
{[}layer 6{]} & fc6 (1); sigmoid                                                                                              \\ \hline
\end{tabular}
}
\end{table}

\subsection{Saliency Object Detection}
The network for saliency detection shown in the Fig. \ref{fig:workflow} is the extended version of the SalGAN \cite{pan2017salgan}. The network is optimized for three different losses naming, content loss i.e., binary cross entropy (BCE), adversarial loss and cyclic consistency loss. Cyclic consistency loss is introduced to limit the space of possible mapping function. The three losses can be formulated as,
\begin{equation}
L_{BCE}^{SOD}=-\frac{1}{n}\sum_{i=1}^{N} (z_i.Log\,(z_i^{p})+(1-z_i).Log\,(1-z_i^{p})) 
\end{equation}
\begin{equation}
L_{Adversarial}^{SOD}=\frac{1}{n}\sum_{i=1}^{N} -Log\,D2(G2(G1(x_i)))
\end{equation}
\begin{equation}
L_{Cyclic}^{SOD}=\frac{1}{n}\sum_{i=1}^{N}\parallel G3(G2(G1(x_i))) - G1(x_i)) \parallel_2 
\end{equation}
The overall joint optimization objective of training the network can be formulated as,
\begin{equation}
L^{SOD}=L_{BCE}^{SOD}+w_{2}.L_{Adversial}^{SOD}+w_{3}.L_{Cyclic}^{SOD}
\end{equation}

 The specification of discriminator \cite{DBLP:journals/corr/MaoSY16a} architecture used for denoising and salient object detection is given in  Table. \ref{tab:tab1} respectively. The generator \cite{DBLP:journals/corr/MaoSY16a} architecture used for salient object detection is  given in Table. \ref{tab:tab2}. The architecture given in Fig. \ref{fig:rednet} is also used for generator3 (G3) to learn reverse mapping from saliency map to corresponding clean image.
 
 \begin{table}[]
\centering
\caption{Generator Network Architecture used for Salient Object Detection. s and p denote stride and padding respectively.}
\label{tab:tab2}
\scalebox{0.52}{
\begin{tabular}{@{}ll@{}}
\toprule
layers         & Image                                                                                                                                                \\ \midrule
{[}layer 1{]}  & \begin{tabular}[c]{@{}l@{}}conv1\_a (1,1,64), s=1, p=1; ReLU\\ conv1\_b (3,3,64), s=1, p=1; ReLU\end{tabular}                                        \\ \midrule
              & pool1 (2,2), s=2, p=0                                                                                                                                \\ \midrule
{[}layer 2{]}  & \begin{tabular}[c]{@{}l@{}}conv2\_a (3,3,128), s=1, p=1; ReLU\\ conv2\_b (3,3,128), s=1, p=1; ReLU\end{tabular}                                      \\ \midrule
              & pool2 (2,2), s=2, p=0                                                                                                                                \\ \midrule
{[}layer 3{]}  & \begin{tabular}[c]{@{}l@{}}conv3\_a (3,3,256), s=1, p=1; ReLU\\ conv3\_b (3,3,256), s=1, p=1; ReLU\\ conv3\_c (3,3,256), s=1, p=1; ReLU\end{tabular} \\ \midrule
              & pool3 (2,2), s=2, p=0                                                                                                                                \\ \midrule
{[}layer 4{]}  & \begin{tabular}[c]{@{}l@{}}conv4\_a (3,3,512), s=1, p=1; ReLU\\ conv4\_b (3,3,512), s=1, p=1; ReLU\\ conv4\_c (3,3,512), s=1, p=1; ReLU\end{tabular} \\ \midrule
              & pool4 (2,2), s=2, p=0                                                                                                                                \\ \midrule
{[}layer 5{]}  & \begin{tabular}[c]{@{}l@{}}conv5\_a (3,3,512), s=1, p=1; ReLU\\ conv5\_b (3,3,512), s=1, p=1; ReLU\\ conv5\_c (3,3,512), s=1, p=1; ReLU\end{tabular} \\ \midrule
{[}layer 6{]}  & \begin{tabular}[c]{@{}l@{}}conv6\_a (3,3,512), s=1, p=1; ReLU\\ conv6\_b (3,3,512), s=1, p=1; ReLU\\ conv6\_c (3,3,512), s=1, p=1; ReLU\end{tabular} \\ \midrule
              & upsample6 (2,2), s=2, p=0                                                                                                                            \\ \midrule
{[}layer 7{]}  & \begin{tabular}[c]{@{}l@{}}conv7\_a (3,3,512), s=1, p=1; ReLU\\ conv7\_b (3,3,512), s=1, p=1; ReLU\\ conv7\_c (3,3,512), s=1, p=1; ReLU\end{tabular} \\ \midrule
              & upsample7 (2,2), s=2, p=0                                                                                                                            \\ \midrule
{[}layer 8{]}  & \begin{tabular}[c]{@{}l@{}}conv8\_a (3,3,256), s=1, p=1; ReLU\\ conv8\_b (3,3,256), s=1, p=1; ReLU\\ conv8\_c (3,3,256), s=1, p=1; ReLU\end{tabular} \\ \midrule
              & upsample8 (2,2), s=2, p=0                                                                                                                            \\ \midrule
{[}layer 9{]}  & \begin{tabular}[c]{@{}l@{}}conv9\_a (3,3,128), s=1, p=1; ReLU\\ conv9\_b (3,3,128), s=1, p=1; ReLU\end{tabular}                                      \\ \midrule
              & upsample9 (2,2), s=2, p=0                                                                                                                            \\ \midrule
{[}layer 10{]} & \begin{tabular}[c]{@{}l@{}}conv10\_a (3,3,64), s=1, p=1; ReLU\\ conv10\_b (3,3,64), s=1, p=1; ReLU\end{tabular}                                      \\ \midrule
              & output (1,1,1), s=1, p=0; Sigmoid                                                                                                                    \\ \bottomrule
\end{tabular}
}
\end{table}

\subsection{Joint Optimisation of Denoising and Saliency Prediction}

We initialise end-to-end coupled network i.e., DSAL-GAN for joint optimization of denoising and saliency prediction by taking pre-trained weights of network1 and network2 as given in Fig. \ref{fig:workflow}. We consider pre-trained weights as initial weights to finetune the combined network into an end-to-end manner.


\section{Experiments and Results}\label{sec:results}

\subsection{Datasets}
\label{sec:Results}
We have trained our model on benchmark datasets such as SOD \cite{movahedi2010design}, MSRA 10k \cite{ChengPAMI} and ECSSD \cite{yan2013hierarchical}. We have created a synthetic dataset using the input images from these datasets and degraded them using Gaussian noise with variance ranging from [10, 30, 50, 80]. The experimental results are obtained on these synthesized datasets for various baseline saliency detection models.
\vspace{-0.1in}

\subsection{Results and Discussion}
\label{sec:Results}
\begin{figure}[thpb]
       \centering
       \fbox{
       \includegraphics[scale=0.35]{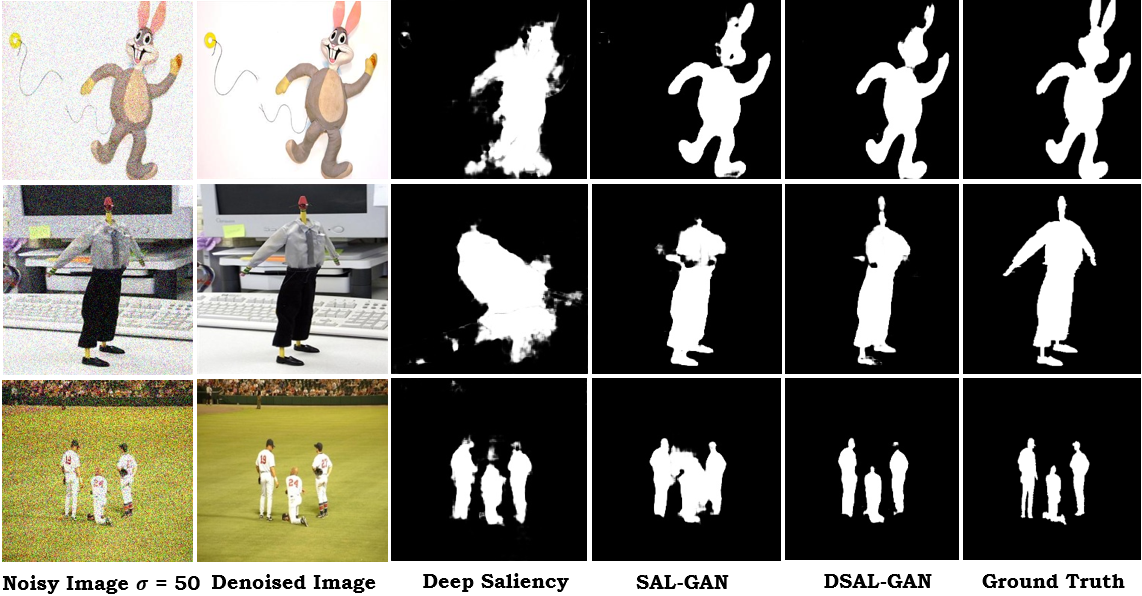}}
       \caption{From left to right, a) Noisy input image, b) Denoised image, c) Deep saliency \cite{DBLP:journals/corr/LiZWYWZLW15}, d) Sal-GAN \cite{DBLP:journals/corr/MaoSY16a}, e) Proposed DSAL-GAN, f) Ground Truth.}
       \label{fig:results}
\end{figure}

The comparison of DSAL-GAN with other state-of-the-art saliency detection framework is shown in Fig. \ref{fig:results} (\textit{for illustration purposes we have shown results for $\sigma$=50 only}).  The performance of the proposed framework shows performance gain over the other state of the art saliency detection frameworks for noisy images. Table \ref{tab:tab3} represents the comparative analysis between these techniques over several performance evaluation metrics average F-measure (aveF), maximum F-measure (maxF), Area under the curve (AUC), Mean Average Error (MAE). The performance of the proposed framework shows far more effective detection of salient regions as compared to the state of the art saliency detection frameworks for noisy images. The proposed framework provides an improvement of $\sim 10\%$ and $16\%$ in aveF, $\sim 9.5\%$ and $16\%$ in maxF, and $\sim 9.6\%$ and $21\%$ in AUC and a drop of $\sim 38\%$ and $47\%$ in MAE with respect to Sal-GAN and Deep Saliency respectively for images having noise variance, $\sigma= 50$. Performance gain in AUC represents the increase in the pixel classification accuracy. It can be evidently observed that the proposed framework  outperforms various state of the art  saliency detection frameworks on noise induced synthetic dataset. The experimental results validates that there is decrease in AUC with increase in noise as shown in Tab. \ref{tab:tab4}. 

\textit{Baseline}: We have benchmarked the performance of DSAL-GAN against the state-of-the-art frameworks like Sal-GAN and Deep Saliency on our synthetic dataset. We observe a performance drop in F1-score, AUC, MAE  metrics for Sal-GAN and Deep Saliency, when they are trained on our synthetic dataset.

\begin{table}[]
\centering
\caption{Comparison of algorithms on different benchmark datasets for $\sigma$=50.}
\label{tab:tab3}
\vspace{0.1in}
\begin{tabular}{lllll}
\hline
Algorithms &  & MSRA-10K & ECSSD & SOD \\ \hline
DSAL-GAN & \begin{tabular}[c]{@{}l@{}}aveF $\uparrow$\\ maxF $\uparrow$\\ AUC $\uparrow$\\ MAE $\downarrow$\end{tabular} & \begin{tabular}[c]{@{}l@{}}0.6523\\ 0.7343\\ 0.9012\\ 0.0923\end{tabular} & \begin{tabular}[c]{@{}l@{}}0.6328\\ 0.7235\\ 0.8503\\ 0.1382\end{tabular} & \begin{tabular}[c]{@{}l@{}}0.6019\\ 0.7104\\ 0.8309\\ 0.1611\end{tabular} \\ \hline
SalGAN & \begin{tabular}[c]{@{}l@{}}aveF $\uparrow$\\ maxF $\uparrow$\\ AUC $\uparrow$\\ MAE $\downarrow$\end{tabular} & \begin{tabular}[c]{@{}l@{}}0.6008\\ 0.7123\\ 0.8221\\ 0.1699\end{tabular} & \begin{tabular}[c]{@{}l@{}}0.5608\\ 0.6431\\ 0.7818\\ 0.2308\end{tabular} & \begin{tabular}[c]{@{}l@{}}0.5493\\ 0.6218\\ 0.7523\\ 0.2407\end{tabular} \\ \hline
Deep Saliency & \begin{tabular}[c]{@{}l@{}}aveF $\uparrow$\\ maxF $\uparrow$\\ AUC $\uparrow$\\ MAE $\downarrow$\end{tabular} & \begin{tabular}[c]{@{}l@{}}0.5818\\ 0.7010\\ 0.6523\\ 0.2023\end{tabular} & \begin{tabular}[c]{@{}l@{}}0.5100\\ 0.5923\\ 0.7404\\ 0.2646\end{tabular} & \begin{tabular}[c]{@{}l@{}}0.5004\\ 0.5757\\ 0.7308\\ 0.2728\end{tabular} \\ \hline
\end{tabular}
\end{table}

\begin{table}[]
\centering
\caption{AUC for different standard deviation ($\sigma$) values on benchmark datasets.}
\label{tab:tab4}
\vspace{0.1in}
\begin{tabular}{@{}lllll@{}}
\toprule
Standard Deviation & $\sigma$=10 & $\sigma$=30 & $\sigma$=50 & $\sigma$=80 \\ \midrule
MSRA-10k           & 0.939    & 0.921    & 0.901    & 0.4512   \\ \midrule
ECSSD              & 0.9006   & 0.8714   & 0.8503   & 0.4163   \\ \midrule
SOD                & 0.8814   & 0.8627   & 0.8309   & 0.3942   \\ \bottomrule
\end{tabular}
\end{table}



\vspace{-0.2in}
\section{Conclusion}
We have performed saliency detection on the noisy images using two generative-adversarial networks trained end-to-end. The denoising network utilized skip connection based convolutional network as a generative framework whereas the saliency detection network used a simple encode-decoder based convolutional network as a generative framework. The saliency detection network was optimized with a combination of three losses namely: content loss, adversarial loss and cycle consistency loss. The denoising network was optimized on content loss and adversarial loss. The proposed network performed reasonably well in comparison to other state-of-the-art saliency detection methods. We have also demonstrated that the use of cycle-consistency loss while training the saliency detection network has enhanced the results to a great extent.
\label{sec:conclusion}

\vspace{-0.20in}

\bibliographystyle{IEEEbib}
\bibliography{refs}

\begin{thebibliography}{10}

\bibitem{park2017transfer}
Ji~Hwan Park, Ievgeniia Gutenko, and Arie~E Kaufman,
\newblock ``Transfer function-guided saliency-aware compression for
  transmitting volumetric data,''
\newblock {\em IEEE Transactions on Multimedia}, 2017.

\bibitem{li2018closed}
Shengxi Li, Mai Xu, Yun Ren, and Zulin Wang,
\newblock ``Closed-form optimization on saliency-guided image compression for
  hevc-msp,''
\newblock {\em IEEE Transactions on Multimedia}, vol. 20, no. 1, pp. 155--170,
  2018.

\bibitem{mademlis2018regularized}
Ioannis Mademlis, Anastasios Tefas, and Ioannis Pitas,
\newblock ``Regularized svd-based video frame saliency for unsupervised
  activity video summarization,''
\newblock in {\em 2018 IEEE International Conference on Acoustics, Speech and
  Signal Processing (ICASSP)}. IEEE, 2018, pp. 2691--2695.

\bibitem{zhou2018perceptually}
Yinzuo Zhou, Luming Zhang, Chao Zhang, Ping Li, and Xuelong Li,
\newblock ``Perceptually aware image retargeting for mobile devices,''
\newblock {\em IEEE Transactions on Image Processing}, vol. 27, no. 5, pp.
  2301--2313, 2018.

\bibitem{pan2017salgan}
Junting Pan, Cristian~Canton Ferrer, Kevin McGuinness, Noel~E O'Connor, Jordi
  Torres, Elisa Sayrol, and Xavier Giro-i Nieto,
\newblock ``Salgan: Visual saliency prediction with generative adversarial
  networks,''
\newblock {\em arXiv preprint arXiv:1701.01081}, 2017.

\bibitem{DBLP:journals/corr/MaoSY16a}
Xiao{-}Jiao Mao, Chunhua Shen, and Yu{-}Bin Yang,
\newblock ``Image restoration using convolutional auto-encoders with symmetric
  skip connections,''
\newblock {\em CoRR}, vol. abs/1606.08921, 2016.

\bibitem{movahedi2010design}
Vida Movahedi and James~H Elder,
\newblock ``Design and perceptual validation of performance measures for
  salient object segmentation,''
\newblock in {\em Computer Vision and Pattern Recognition Workshops (CVPRW),
  2010 IEEE Computer Society Conference on}. IEEE, 2010, pp. 49--56.

\bibitem{ChengPAMI}
Ming-Ming Cheng, Niloy~J. Mitra, Xiaolei Huang, Philip H.~S. Torr, and Shi-Min
  Hu,
\newblock ``Global contrast based salient region detection,''
\newblock {\em IEEE TPAMI}, vol. 37, no. 3, pp. 569--582, 2015.

\bibitem{yan2013hierarchical}
Qiong Yan, Li~Xu, Jianping Shi, and Jiaya Jia,
\newblock ``Hierarchical saliency detection,''
\newblock in {\em Proceedings of the IEEE Conference on Computer Vision and
  Pattern Recognition}, 2013, pp. 1155--1162.

\bibitem{DBLP:journals/corr/LiZWYWZLW15}
Xi~Li, Liming Zhao, Lina Wei, Ming{-}Hsuan Yang, Fei Wu, Yueting Zhuang, Haibin
  Ling, and Jingdong Wang,
\newblock ``Deepsaliency: Multi-task deep neural network model for salient
  object detection,''
\newblock {\em CoRR}, vol. abs/1510.05484, 2015.

\end{thebibliography}

\end{document}